\documentclass[10pt,twocolumn,letterpaper]{article}

\usepackage{iccv}
\usepackage{times}
\usepackage{epsfig}
\usepackage{graphicx}
\usepackage{amsmath}
\usepackage{amssymb}

\usepackage{booktabs}
\usepackage{multirow}
\usepackage{xcolor}
\usepackage{arydshln}
\usepackage{wrapfig}
\usepackage{enumitem}
\usepackage[font=small]{caption}
\usepackage{tabularx} 
\usepackage{microtype}
\usepackage{pifont}

\definecolor{newlightblue}{RGB}{0,75,255}
\usepackage[pagebackref=true,breaklinks=true,colorlinks,bookmarks=false,urlcolor={newlightblue},citecolor={newlightblue}]{hyperref}

\iccvfinalcopy %

\usepackage[capitalize]{cleveref}
\crefname{section}{Sec.}{Secs.}
\Crefname{section}{Section}{Sections}
\Crefname{table}{Table}{Tables}
\crefname{table}{Tab.}{Tabs.}

\definecolor{darkgreen}{RGB}{0, 200, 50}

\DeclareMathOperator*{\argmin}{arg\,min}

\newcommand{\bv}[0]{{\mathbf v}}
\newcommand{\ba}[0]{{\mathbf a}}
\newcommand{\bA}[0]{{\mathbf A}}
\newcommand{\br}[0]{{\mathbf r}}

\newcommand{\cmark}{\ding{51}}

\newcommand{\supparxiv}[2]{#2}
\newcommand{\mypar}[1]{\vspace{-3mm}\paragraph{#1}}
\def\upvspacefig{\vspace{0mm}}
\def\iccvPaperID{1105} %

\ificcvfinal\fi

\begin{document}

\title{Sound Localization from Motion:\\ Jointly Learning Sound Direction and Camera Rotation}

\author{Ziyang Chen \qquad Shengyi Qian \qquad Andrew Owens \vspace{3.5mm}\\ 
University of Michigan~~~~\\
}

\maketitle

\begin{abstract}

The images and sounds that we perceive undergo subtle but geometrically consistent changes as we rotate our heads. In this paper, we use these cues to solve a problem we call Sound Localization from Motion~(SLfM): jointly estimating camera rotation and localizing sound sources. We learn to solve these tasks solely through self-supervision. A visual model predicts camera rotation from a pair of images, while an audio model predicts the direction of sound sources from binaural sounds. We train these models to generate predictions that agree with one another. At test time, the models can be deployed independently. To obtain a feature representation that is well-suited to solving this challenging problem, we also propose a method for learning an audio-visual representation through {\em cross-view binauralization}: estimating binaural sound from one view, given images and sound from another.
Our model can successfully estimate accurate rotations on both real and synthetic scenes, and localize sound sources with accuracy competitive with state-of-the-art self-supervised approaches. Project site: \small{\url{https://ificl.github.io/SLfM}}.

\end{abstract}

\section{Introduction}
\label{sec:intro}

\begin{figure}[t!]

\centering

\includegraphics[width=\linewidth]{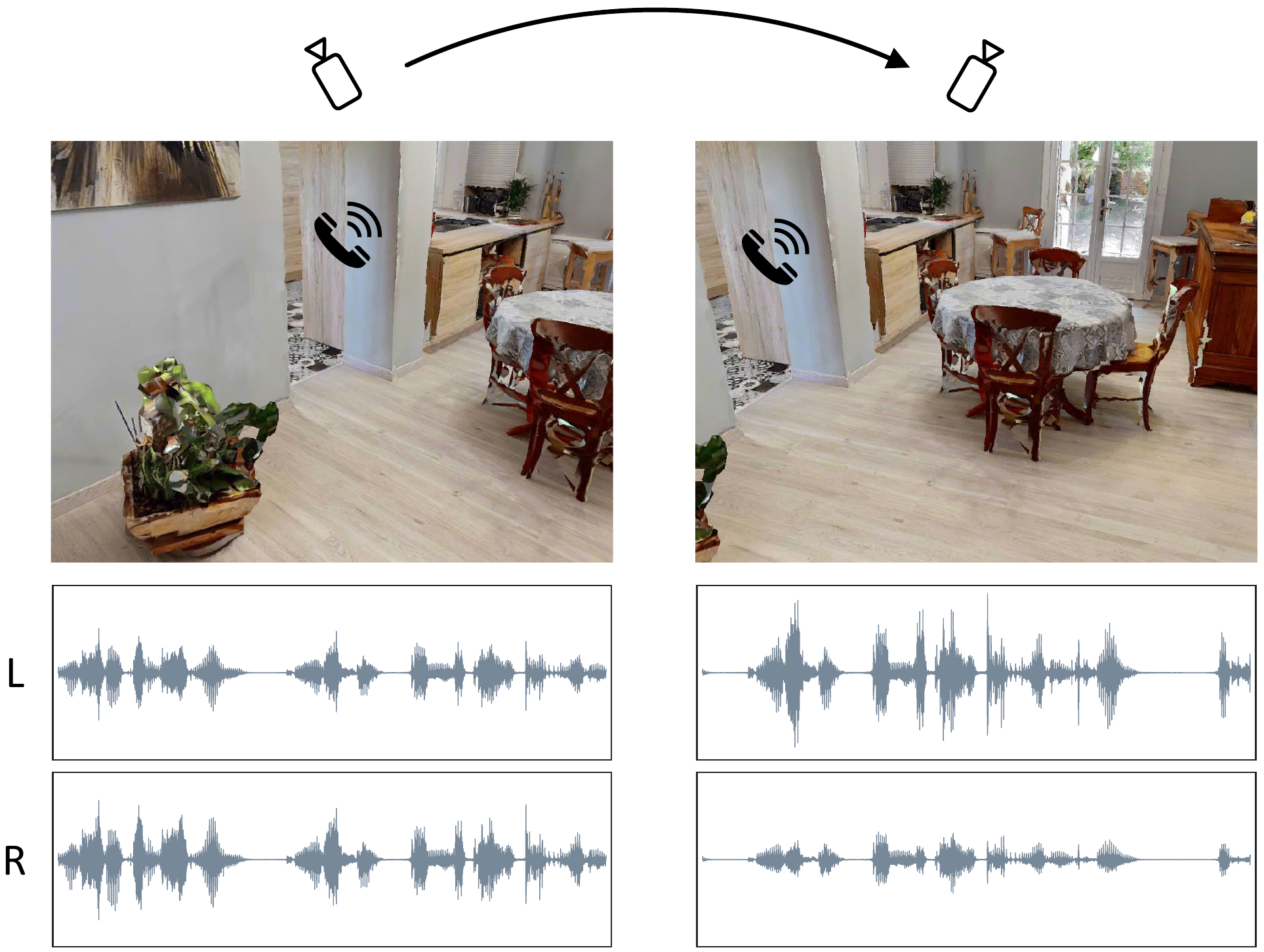}
    
\caption{
{\bf Images and sounds change in geometrically consistent ways.} 
For example, when we rotate to the right, a sound source that is initially in front of us becomes louder in our left ear. We use these cues to jointly train models for two tasks: localizing sounds from binaural audio and estimating camera rotation from images. The two models are trained entirely through self-supervision, by learning to produce outputs that agree with one other.
}

 \supparxiv{\vspace{-1em}}{}

\label{fig:teaser}
\end{figure}

As you rotate your head, the images and sounds that you perceive change in  geometrically consistent ways. For example, after turning to the right, a sound source that was directly in front of you will become louder in your left ear and quieter in your right, while simultaneously the visual scene will move right-to-left across your visual field (\cref{fig:teaser}).

We hypothesize that these co-occurring audio and visual signals provide ``free'' supervision that captures geometry, including the motion made by a camera and the direction of sound sources. These  are each core problems in machine perception, but are largely studied separately, often using supervised methods that rely on difficult-to-acquire labeled training data, such as annotated sound directions. We take inspiration from self-supervised approaches to structure from motion~\cite{zhou2017unsupervised}, which learn to estimate 3D structure and camera pose by solving both tasks simultaneously.

Analogously, we propose a problem we call {\em sound localization from motion} (SLfM): jointly estimating camera rotation from images and the sound direction from binaural audio. By solving both tasks simultaneously, we avoid the need for labeled training data. Our models provide each other with self-supervision: a visual model predicts the rotation angle between pairs of images, while an audio model predicts the azimuth of sound sources. We force their predictions to agree with one another, such that  changes in rotation are consistent with changes in sound direction and binaural cues. After training, the models can be deployed independently, without multimodal data at test time.

This is a challenging task that requires perceiving motion in images and binaural cues in audio. Our second contribution is a method for learning representations that are well-suited to this task through {\em cross-view binauralization}. We train a network to convert mono to binaural sound for one viewpoint, given an audio-visual pair sampled from another viewpoint. Since the sound source is not necessarily visible in the images, the only way to successfully solve this pretext task is by analyzing the changes in the camera pose and predicting how they affect the sound direction. 

All components of our model are {entirely self-supervised} and are trained solely on unlabeled audio-visual data. Our results suggest that paired audio-visual data provides a useful and complementary signal for
learning about geometry. In contrast to other audio or visual self-supervised pose estimation methods, we obtain supervision from abundantly available audio data, thus avoiding the need of 3D ground truth or correspondences between pixels~\cite{zhou2017unsupervised,zou2020learning} or audio samples~\cite{chen2022sound}. %
Through experiments, we show:
\begin{itemize}[leftmargin=*,topsep=1pt, noitemsep]
\item Paired audio-visual data provides a supervisory signal for pose estimation tasks. 
\item We obtain competitive performance with state-of-the-art self-supervised sound localization methods~\cite{chen2022sound}.
\item We obtain strong rotation estimation performance, and our model generalizes to Stanford2D3D~\cite{armeni2017joint} dataset, where it is competitive with classic sparse feature matching methods.
\item The features we learn through our pretext task outperform other representations for our downstream tasks.
\end{itemize}

\section{Related Work}
\label{sec:related_work}
\vspace{3mm}

\mypar{Audio for spatial perception.} 
Recent works have explored the use of sound for spatial understanding.
Purushwalkam \etal~\cite{purushwalkam2021audio} reconstructed floor plans in simulated environments~\cite{chen2020soundspaces}. 
Chen \etal~\cite{chen2021structure} used ambient sounds from environments to learn about scene structures. 
Konno \etal~\cite{konnoaudio} integrated sound localization to visual SfM while do not jointly learn them. 
Other work learns representations for spatial audio-visual tasks. Yang \etal~\cite{yang2020telling} predicted whether stereo channels are swapped in a video, and Morgado \etal~\cite{morgado2020learning} solved a spatial alignment task. The learned representations are then used to improve localization, up-mixing, and segmentation models. In contrast, we learn camera pose and sound localization solely from self-supervision, obtaining angular predictions without labeled data. 
Other work uses echolocation sounds to learn representations~\cite{gao2020visualechoes,yang2022camera} and predict depth maps~\cite{christensen2020batvision,parida2021beyond} and estimate camera poses~\cite{yang2022camera} using labeled data.
In contrast, our proposed approach jointly learns binaural sound localization and camera pose through passive audio sensing, without supervision.

\begin{figure*}[t]
    \centering
    \upvspacefig
    \includegraphics[width=\linewidth]{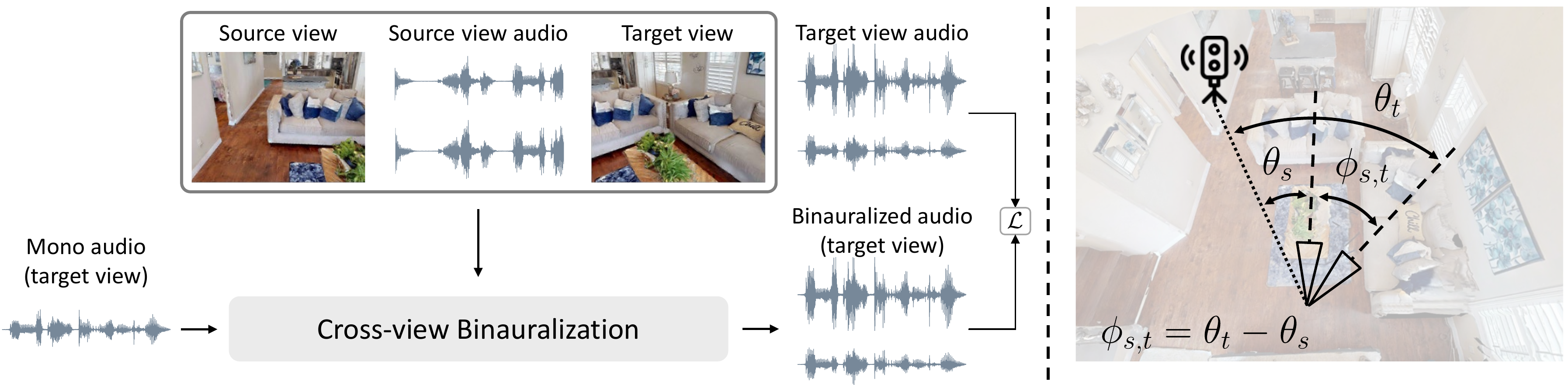}

    \begin{flushleft}
        \vspace{-4mm}
        \hspace{30mm} (a) Cross-view audio prediction \hspace{44.5mm} (b) Sound localization from motion
         \vspace{-3mm}
    \end{flushleft}    
    \caption{{\bf Method overview.} (a) We learn a feature representation by predicting how changes in images lead to changes in sound in a cross-view binauralization pretext task. We convert mono sound to binaural sound at a target viewpoint, after conditioning the model on observations from a source viewpoint. (b) We use the representation to jointly solve two pose estimation tasks: visual rotation estimation and binaural sound localization. We train the visual rotation angle, $\phi_{s,t}$, to be consistent with the difference in predicted sound angles $\theta_s$ and $\theta_t$. } 
    \label{fig:method}
\end{figure*}

\mypar{Acoustic synthesis and spatialization.} 
Researchers have explored visually-guided sound synthesis~\cite{gan2020foley,ghose2020autofoley,iashin2021taming,du2023conditional} and text-guided audio synthesis~\cite{kreuk2022audiogen,yang2022diffsound,huang2023make}. Additionally, researchers have investigated generating realistic environmental acoustics using visual information~\cite{chen2022visual,singh2021image2reverb,chen2022soundspaces,majumder2022few}.
Chen \etal~\cite{chen2023novel} introduced the novel-view acoustic synthesis task, which synthesizes binaural sound at the target view using audio and visual information from a source view. Liang \etal~\cite{liang2023av} proposed an audio-visual neural field in real-world audio-visual scenes.
Many recent works have proposed to generate spatial audio from mono audio using visual cues~\cite{morgado2018self,gao20192,rachavarapu2021localize,xu2021visually,lin2021exploiting,zhou2020sep,garg2021geometry}, or the relative pose between sound sources and the receiver~\cite{richard2021neural,huang2022end}. Inspired by these works, our feature learning approach learns spatial representations through an audio prediction task. 

\mypar{Binaural sound localization.} Humans have the ability to localize sound sources from binaural sound~\cite{rayleigh1907perception}.  Traditional approaches estimate interaural time delays via cross-correlation using hand-crafted features~\cite{knapp1976Generalized}, factorization methods~\cite{schmidt1986multiple}, or loudness differences between ears~\cite{rayleigh1907xii,wang2006computational}. 
Chen \etal~\cite{chen2022sound} adapted methods from self-supervised visual tracking to the problem of binaural sound localization. Similarly, we estimate direction through self-supervision. However, we obtain our supervision through cross-modal supervision from vision instead of from correspondence cues. Moreover, we also obtain visual camera rotation estimation through our learning process. Francl~\cite{francl2022modeling} learned representations of sound location with a contrastive loss where positive and negative examples are selected based on the extent of head movements. Other works have used supervised learning techniques with labeled data to localize sound sources in reverberant environments~\cite{adavanne2018direction,vecchiotti2019end,yalta2017sound}. 
Unlike these methods, our model learns 3D sound localization without labels.

\mypar{Camera pose estimation.}
Traditional methods for camera pose estimation are based on finding correspondences between images and then solving an optimization problem using constraints from multi-view geometry~\cite{Hartley04}. These include structure from motion methods that estimate full  pose~\cite{schonberger2016structure} and camera rotation~\cite{brown2007automatic}.
Recent methods have directly predicted camera pose using neural networks, including methods that use photos~\cite{qian2020associative3d,kendall2015posenet,melekhov2017relative,jin2022perspective,ma2022virtual} or RGB-D scans~\cite{yang2020extreme,yang2019extreme,banani2021unsupervisedr,banani2021bootstrap}. 
Our setup is similar to work that learns relative camera poses from sparse views~\cite{jin2021planar,cai2021extreme}, and we use their network architectures. However, we learn the camera pose through cross-modal supervision from audio, rather than from labels.  Our work is also closely related to methods that learn structure from motion through self-supervision~\cite{zhou2017unsupervised,zou2020learning}, such as by jointly learning models that perform depth and camera pose estimation with photoconsistency constraints. In contrast, our visual model's learning signal comes solely from audio-based supervision, and we jointly learn audio localization.

\mypar{Audio-visual learning.} 
A number of works have focused on learning multimodal representations for audio and vision, taking into account semantic correspondence and temporal synchronization~\cite{owens2018audio, xiao2020audiovisual,asano2020labelling,morgado2021audio,owens2016visually,afouras2022self,mittal2022learning}. 
Other approaches study audio-visual sound localization~\cite{hu2022mix,mo2022closer,chen2021localizing,mo2022localizing}, source separation~\cite{majumder2021move2hear,gao2021visualvoice,majumder2022active,tzinis2022audioscopev2}, active speaker detection~\cite{afouras2020self,tao2021someone,alcazar2021maas},  navigation~\cite{chen2020soundspaces,chen2020learning,chen2021semantic} and forensics~\cite{zhou2021joint,haliassos2022leveraging,feng2023self}. Our focus, in contrast, is on utilizing multi-view audio-visual signals to learn geometry.

\section{Sound Localization from Motion}
\label{sec:method}

We address the {\em sound localization from motion}~(SLfM) task: predicting the azimuth of a sound source from binaural audio and camera rotation from two images. First, we present a self-supervised representation that can be used to solve this downstream task. Then, we show how the representation can be used to solve the task.

\subsection{Learning representation via spatialization}

We learn an audio-visual representation that conveys spatial information by solving a {\em cross-view binauralization} task: 
converting mono sound to stereo for one viewpoint, given an audio-visual pair sampled from another viewpoint~(\cref{fig:method}a). In order to successfully solve the task, a model must implicitly estimate the sound direction in the source view and predict how the change in viewpoint will affect the sound in the target view. A key difference between this task and traditional, single-view binauralization~\cite{gao20192} is that the sound source {\em need not be visible} in any of the images. Hence, the task cannot be solved from the target audio-visual pair alone. 

We binauralize the sound $\ba_t$ at the target view through an audio predictor $\mathcal{F}_{\theta}$. To make our representation suitable for downstream tasks, we factorize it into visual features $f_v(\bv_s, \bv_t)$, which are intended to create features relevant to relative pose, and audio features $f_a(\ba_s)$, which capture sound localization cues. We predict the binauralized audio $\hat{\ba}_t$ at the target view from mono audio $\bar{\ba}_t$, the visual change of $(\bv_s, \bv_t)$ and binaural sound $\ba_s$ heard at the source viewpoint: 
\begin{equation}
    \hat{\ba}_t = \mathcal{F}_{\theta}\left(\bar{\ba}_t, f_v(\bv_s, \bv_t), f_a(\ba_s)\right).
    \label{eq:generative}
    \vspace{-1mm}
\end{equation}

We represent audio $\ba$ as a spectrogram $\bA$ using short-time Fourier transform~(STFT). Following Gao \etal~\cite{gao20192}, given the mix of stereo audio $\bar{\bA}_t = \text{STFT}(\ba^L_t + \ba^R_t)$, we predict the difference of two channels $\bA_t = \text{STFT}(\ba^L_t - \ba^R_t)$. We optimize the $L1$ loss between predicted spectrogram $\hat{\bA}_t$ and ground-truth spectrogram $\bA_t$:
  \vspace{-2mm}
\begin{equation}
     \mathcal{L}_{\mathrm{pretext}} = ||\hat{\bA}_t - \bA_t||_1. \\
  \label{eq:loss}
  \vspace{-2mm}
\end{equation}

\mypar{Multi-view binauralization.} 
Following Zhou~\etal~\cite{zhou2017unsupervised}, we improve our representations by binauralizing sounds at $N$ different target viewpoints, using observations from a single source viewpoint $s$.  We hypothesize that  jointly solving spatialization problems from a single viewpoint for multiple target viewpoints would require the model to make more accurate predictions of sound source locations, thereby improving the estimation of view changes:
\begin{equation}
    \mathcal{L}_{\mathrm{pretext}}  = \frac{1}{N} \sum_i^{N} ||\mathcal{F}_{\theta}\left(\bar{\bA}_{i}, f_v(\bv_s, \bv_{i}), f_a(\ba_s)\right) - \bA_{i}||_1.
 \label{eq:multi_binaural}  
 \vspace{-3mm}
\end{equation}

\subsection{Estimating pose and localizing sounds}

We now address the problem of learning models for sound localization and pose estimation, using our self-supervised audio-visual features. Given two views, we predict sound directions and relative rotation. We train the model to make these two predictions consistent with one another, while using simple binaural constraints to resolve ambiguities. 

We are given images $\bv_s$ and $\bv_t$ (rotated views recorded at the same position) and learned visual embedding $f_v$. We predict the scalar rotation angle $\phi_{s, t}$ via the encoder $g_v$: 
\begin{equation}
    \phi_{s, t} = g_v\left( f_v(\bv_s, 
    \bv_t)\right), \hspace{1mm} R_{s, t}=
    \begin{bmatrix}
        \cos \phi_{s, t} & - \sin \phi_{s, t} \\
        \sin \phi_{s,t} & \cos\phi_{s, t}
    \end{bmatrix},
\end{equation}
where $R_{s, t}$ is 2D rotation matrix of $\phi_{s, t}$.
Following common practice in indoor scene reconstruction, we give the camera a fixed downward tilt~\cite{zhi2021place,qian2021recognizing,zhang2017physically} and only estimate azimuth~\cite{jin2021planar}. This is also a common assumption in audio localization~\cite{adavanne2018direction,vecchiotti2019end}, since azimuth has strong binaural cues.

Similarly, we predict the azimuths of the sound sources using audio features  and  the encoder $g_a$: 
\begin{equation}
    \theta_{i} = g_a\left( f_a(\ba_i)\right), \quad \br_i = \begin{bmatrix}
        \cos \theta_{i} & \sin \theta_{i}
    \end{bmatrix}^\mathsf{T},
\end{equation}
where we represent the azimuth as a vector $\br_i$

\mypar{Cross-modal geometric consistency.}
When the camera is rotated, the sound source ought to rotate in the opposite direction~(\cref{fig:method}b). For example, a $30^\circ$ clockwise camera rotation should result in a $30^\circ$ counterclockwise rotation in sound direction.  Such a constraint could be converted into a loss:
\begin{equation}
    \mathcal{L}_{\mathrm{rot}} = \lVert\br_{s} -  R_{s, t}\br_{t}\rVert^2. %
    \label{eq:geometric}
\end{equation}

However, a well-known ambiguity called front-back confusion~\cite{rayleigh1907xii,fischer2020front} exists in binaural sound perception: one cannot generally tell whether a sound is in front of the view, or behind them.
To address that, we use permutation invariant training~\cite{yu2017permutation}~(PIT), and allow the model to use either the predicted sound direction or its reflection about the $x$ axis without penalty. This results in the loss:
\vspace{-1mm}
\begin{equation}
        \mathcal{L}_{\mathrm{geo}} = \min_{
        \substack{
        \hat{\mathbf{r}}_s \in \{\mathbf{r}_s, Q \mathbf{r}_s\} \\
        \hat{\mathbf{r}}_t \in \{\mathbf{r}_t, Q \mathbf{r}_t\}}}~\lVert\hat{\br}_{s} -  R_{s, t}\hat{\br}_{t}\rVert^2, \vspace{-1mm}
    \label{eq:per_geometric}
\end{equation}
\vspace{-1mm}
where $Q = \begin{bmatrix} 1 & 0 \\ 0 & {-1} \end{bmatrix}$ reflects the sound direction. 

\vspace{3mm}
As a consequence of this ambiguity, there are also two possible solutions for the visual rotation model, since the visual rotation matrices can be mirrored about the $x$ axis. For example, one can create a solution with equal loss by multiplying the rotations and sound directions by $Q$. We discuss this ambiguity in more depth in \cref{sec:pose}.

\mypar{Incorporating binaural observations.} 
Without additional constraints, the solution is ambiguous, and may collapse into a trivial solution (\eg, predicting zero for all three angles).\footnote{Work on self-supervised SfM has similar ambiguities~\cite{zhou2017unsupervised}, and deals with them by adding analogous constraints, such as photometric consistency.} To avoid this, we force the model to agree with a simple  binaural cue based on interaural intensity difference~(IID). We predict whether the sound is to the left or right of the viewer, based on whether it is louder in the left or right microphone: $d = \mathrm{sign}(\log\left\vert \frac{\bA^L}{\bA^R}\right\vert)$, where $\vert \bA \vert$ is the magnitude of the spectrogram $\bA$. 
We perform this left/right test at each timestep in the spectrogram and then pool via majority voting (see \supparxiv{supplementary}{\cref{appendix:implement}} for details).
We penalize predictions that are inconsistent with these ``left or right'' observations: %
\vspace{-2mm}
\begin{equation}
    \mathcal{L}_{\mathrm{binaural}} = \mathcal{L}_{\mathrm{BCE}}\left(\sin\theta_i, d_i\right),
    \label{eq:binaural}
\vspace{-1mm}
\end{equation}
where $\mathcal{L}_{\mathrm{BCE}}$ is binary cross entropy loss. 

\mypar{Encouraging symmetry.} To help regularize the model, we also add symmetry constraints. 
For sound localization, swapping the left and right channels of the audio ought to result in a prediction in the opposite direction since the binaural cues are reversed.
For rotation estimation, the relative pose between images $s$ and $t$ should invert the pose from $t$ to $s$. We encourage both constraints via a loss:
\begin{equation}
    \mathcal{L}_{\mathrm{sym}} = \lvert\theta + \theta_{\mathrm{flip}} \rvert + \lvert\phi_{s, t} + \phi_{t, s}\rvert,
    \label{eq:symmetric}
\end{equation}
where $\theta_{\mathrm{flip}}$ is the prediction of sound angle $\theta$ using audio with swapped audio channels, and $\phi_{s,t}$ and $\phi_{t,s}$ are the predicted rotations between cameras $s$ and $t$.

\mypar{Overall loss.} We combine these constraints to obtain an overall loss:
\vspace{-2mm}
\begin{equation}
    \mathcal{L} = \lambda \mathcal{L}_{\mathrm{geo}} + \mathcal{L}_{\mathrm{binaural}} + \mathcal{L}_{\mathrm{sym}},
    \label{eq:overall}
\end{equation}
\vspace{-1mm}
where $\lambda$ is the weight for the geometric loss.

\section{Experiments}
\label{sec:exp}

We have introduced a self-supervised method to learn camera pose and sound localization from audio-visual data.
In experiments, we first evaluate how well our learned representation captures spatial information.
We then evaluate how well our method learns camera pose and sound localization by comparing it with baselines. %
Finally, we show generalization to indoor panorama images Stanford2D3D~\cite{armeni2017joint} and in-the-wild binaural audio~\cite{chen2022sound}.

\begin{figure}[t!]
\centering
\upvspacefig
\includegraphics[width=\linewidth]{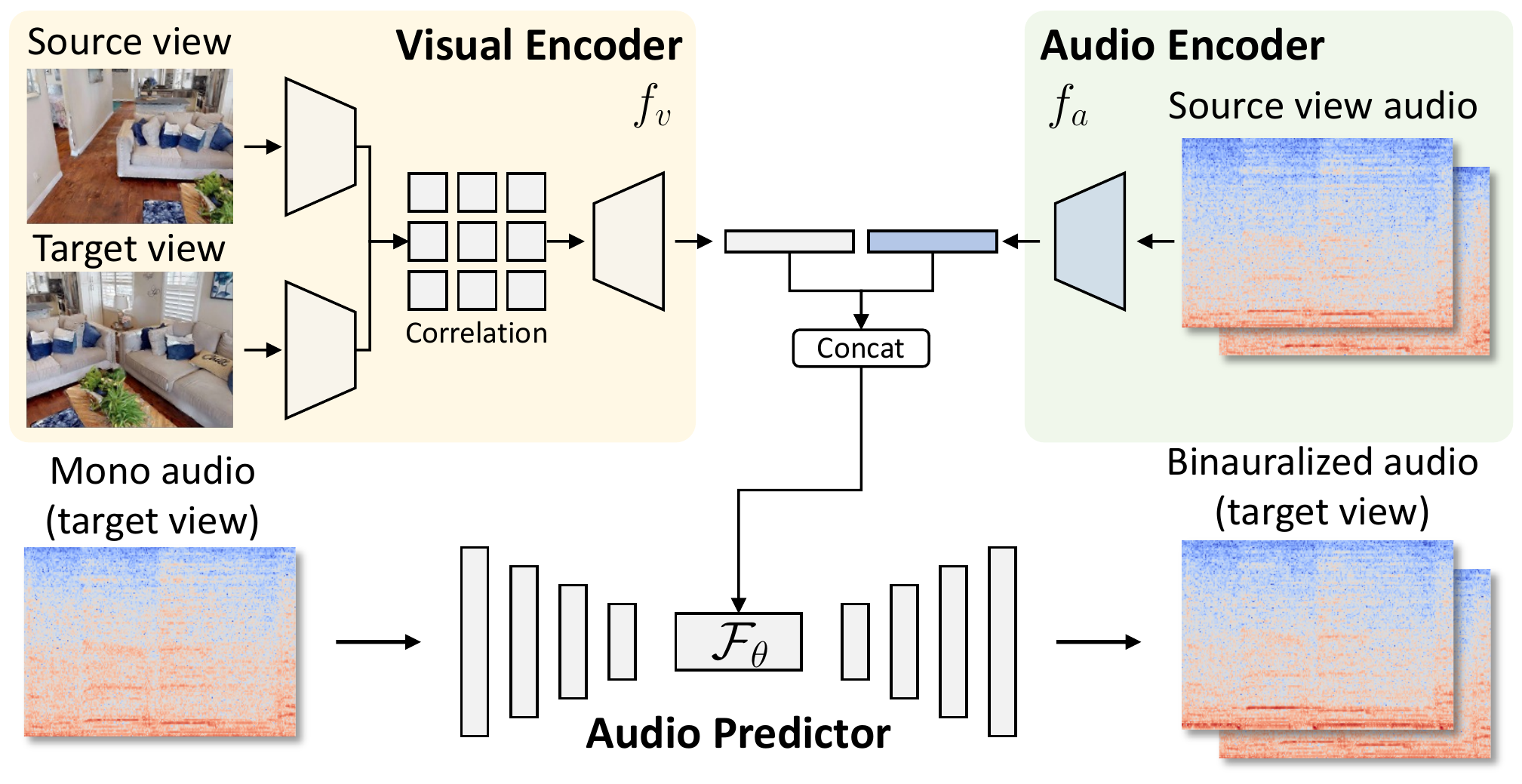}
    
\caption{
{\bf Cross-view binauralization architecture.} We take a mono spectrogram as input and fuse the audio and visual features from the audio and visual encoder respectively to synthesize the binaural spectrogram at the target viewpoint. 
}

\label{fig:arch}
\end{figure}
\subsection{Implementations}
\vspace{3mm}
\mypar{Visual pose encoder.} We follow recent pose estimation work~\cite{cai2021extreme,jin2021planar} and build a Siamese-style visual pose network $f_v$ with ResNet-18~\cite{he2016} as the backbone. We compute dense 4D correlation volumes between the features from the third residual layer and then encode them by convolution layers. We resize images to $320 \times 240$ and encode a pair of images into 512-d features. We use an MLP $g_v$ to map visual features to 1-d logits for our SLfM models. 

\mypar{Binaural audio encoder.} We obtain binaural audio embeddings $f_a(\cdot)$ using ResNet-18~\cite{he2016} that operates on spectrograms. We covert the two-channel waveform of length $L$ to a spectrogram representation of size $256 \times 256 \times 4$ using short-time Fourier transform~(STFT), where we keep both the magnitude and phase of spectrograms. We extract 512-d features of binaural sound with $f_a$ and map them to 1-d logits using an MLP $g_a$.

\mypar{Audio prediction model.} We adopt the light-weighted audio-visual U-Net~\cite{gao20192} to perform binauralization. We feed in spectrograms of size $256 \times 256 \times 2$ and predict the target spectrograms. We concatenate the visual pose features $f_v(\cdot)$ and audio features $f_a(\cdot)$ at the bottleneck of U-Net. We show the architecture of our models in \cref{fig:arch}. Please see \supparxiv{supp.}{\cref{appendix:implement}} for more implementation details.

\subsection{Dataset}
Since there is no public multi-view audio-visual dataset with camera poses and sound direction ground truth, we use the SoundSpaces 2.0 platform~\cite{chen2022soundspaces} to create a dataset. 
Our 3D scenes come from Habitat-Matterport 3D dataset (HM3D)~\cite{ramakrishnan2021hm3d}, which is a large dataset of real 3D scenes.
This setup allows us to have photorealistic images and high-quality spatial audio with real-world acoustics phenomenon (\eg, reverberation), as well as providing the ground-truth camera pose and sound directions that can be used for evaluation. 
We call this dataset HM3D-SS. %

We generate binaural Room Impulse Responses~(RIRs) and images with a $60^\circ$ field of view, using 100 scenes of HM3D~\cite{ramakrishnan2021hm3d}. For each audio-visual example, we randomly place sound sources in the scene with a height range of $(0.7, 1.7)$ meters, and sample 4 different rotated viewpoints at one location within 4 meters. The rotations are limited to $(10^\circ, 90^\circ)$ relative to the source viewpoints. 
We follow the standard practice to set the height to agents to be 1.5m and lock a downward tilt angle~\cite{jin2021planar,zhang2017physically,qian2021recognizing,zhi2021place}.
We render the binaural RIRs and images given the position of agents and sound sources.
We obtain binaural audio by convolving binaural RIRs with mono audio samples from LibriSpeech~\cite{panayotov2015librispeech} and Free Music Archive~\cite{defferrard2016fma}. To ensure that the evaluation tests the model's pose estimation abilities, rather than its ability to visually localize sound sources, the sound sources are not visible on screen. 

We create 50K audio-visual pairs from 200K viewpoints. The audio was rendered with average reverberation of $\text{RT}_{60}=0.4\text{s}$ (see \supparxiv{supp.}{\cref{appendix:implement}} for details).
We divided our data into 81/9/10 scenes for the train/val/test, respectively.

\begin{table}[b!]
\centering
\upvspacefig
\resizebox{0.95\columnwidth}{!}{
\begin{tabular}{clcc}
\toprule
& \multirow{2}{*}{Model} & \multirow{2}{*}{\shortstack[c]{Audio Loc. \\ Acc~(\%)~$\uparrow$}} & \multirow{2}{*}{\shortstack[c]{Camera Rot. \\ Acc~(\%)~$\uparrow$}} \\
&   &  &              \\
\midrule

\parbox[t]{1mm}{\multirow{11}{*}{\rotatebox[origin=c]{90}{\shortstack[c]{LibriSpeech}}}} 
& Random feature &  4.8  & 4.7 \\
& ImageNet~\cite{he2016}+Random &  -- &  56.3  \\
& RotNCE~\cite{francl2022modeling} &  50.9    &   --  \\
& AVSA~\cite{morgado2020learning} &  71.2 &  6.5 \\
\cdashlinelr{2-4}
& Ours--NoA &  --  & 9.6 \\
& Ours--NoV &   53.9   & --\\
& Ours--GTRot &  70.6   & --\\
& Ours--L2R~(3 views) & {\bf 78.5} & 76.1\\
& Ours~(2 views) &  74.5  & 80.0\\
& Ours~(3 views) &  75.4 & {\bf 81.3}\\
\cmidrule(lr){2-4}
& Supervised & 81.5 & 95.8 \\

\bottomrule
\end{tabular}
}

\caption{{\bf Downstream task performance on HM3D-SS dataset.} We report linear probe performance on the audio localization and camera rotation downstream tasks.  }
\label{tab:downstream}
\end{table}

\subsection{Evaluating the learned representation}
First, we directly evaluate the quality of our learned features for rotation estimation and sound localization via linear probing with labeled data (rather than learning them jointly through self-supervision).

\mypar{Baselines and ablations.}
We compare our model with several baselines that use alternative pretext tasks: 1)~{\bf AVSA}~\cite{morgado2020learning}: it learns spatial cues by training a model to spatially align video and audio clips extracted from different viewing angles. We adapt this model to our dataset and train with 4 different views; 2)~{\bf RotNCE}~\cite{francl2022modeling}: it applies contrastive learning on the audio from different angles and uses annotations of the agent's rotation to select positive and negative samples, which results in learning audio spatial representation. For baselines, we use the same architecture for feature extractors to ensure fair comparisons. 

To determine if we utilize visual and audio features from different views to solve the binauralization task, we also study some variants of our models: 
1)~{\bf Ours-NoA}: we only provide visual features for the binauralization task; 
2)~{\bf Ours-NoV}: which only uses audio from the other view to spatialize sounds;
3)~{\bf Ours-GTRot}: we provide ground-truth rotation embedding instead of features from visual frames.

Besides the mono-to-binaural task, we also experiment with another objective: predicting the right channel from the left channel.
We train our {\bf L2R} model with the same setup as our M2B model. (Please see \supparxiv{supp.}{\cref{appendix:additional}} for pretext results.)

\begin{table}[t!]
\small
\centering
\vspace{-1mm}
\upvspacefig
\resizebox{0.9\columnwidth}{!}{
\begin{tabular}{clcc}
\toprule
& \multirow{2}{*}{Model} & \multirow{2}{*}{\shortstack[c]{Audio angle\\MAE~($^\circ$)~$\downarrow$}} & \multirow{2}{*}{\shortstack[c]{Camera angle\\MAE~($^\circ$)~$\downarrow$}} \\
&   & &       \\
\midrule

\parbox[t]{2mm}{\multirow{10}{*}{\rotatebox[origin=c]{90}{\shortstack[c]{LibriSpeech}}}} 
& Chance  &  40.28  & 29.41\\
& SIFT~\cite{lowe2004distinctive} &  -- & 12.2 \\
\cdashlinelr{2-4}
& Ours w/o Reflect &  28.08    & 26.80 \\
& Ours--Prompt &  --  & 9.13\\
& Ours--L2R &  3.22   &  0.80 \\
& Ours &  {\bf 3.17}   & {\bf 0.77} \\

\cmidrule(lr){2-4}
& Ours--Front &  4.48    & 3.08\\
& Ours--GTRot &  1.83  & --\\
\cmidrule(lr){2-4}
& Superglue~\cite{sarlin2020superglue} & -- & 2.47 \\
& Supervised  &  1.71 & 0.46\\

\midrule
\parbox[t]{2mm}{\multirow{9}{*}{\rotatebox[origin=c]{90}{\shortstack[c]{FreeMusic}}}} 
& Chance  &  40.81  & 29.41\\
& SIFT~\cite{lowe2004distinctive} &  -- & 12.2 \\
\cdashlinelr{2-4}
& Ours w/o Reflect &  28.24  & 26.81 \\
& Ours--L2R &  {\bf 3.18}   & {\bf 0.81} \\
& Ours &  3.37   & 0.84 \\
\cmidrule(lr){2-4}
& Ours--Front &  3.99   & 2.49\\
& Ours--GTRot &  1.96  & --\\

\cmidrule(lr){2-4}
& Superglue~\cite{sarlin2020superglue} & -- & 2.47 \\
& Supervised  &  2.50 & 0.46\\

\bottomrule
\end{tabular}
}

\caption{{\bf Sound localization from motion results on HM3D-SS.} We evaluate our SLfM models on each modality independently. }
\label{tab:pose_estimation}
\end{table}

\mypar{Downstream tasks.}
We assess the quality of spatial representations we learned from our pretext tasks in two downstream tasks: relative camera rotation and 3D sound localization. We formulate them as classification problems, where angles are categorized into 64 bins, and we use accuracy as the evaluation metric.
To evaluate the learned features, we freeze them and train a linear classifier on the downstream tasks. We compare the performance of our features with those learned from RotNCE~\cite{francl2022modeling}, AVSA~\cite{morgado2020learning}, ImageNet~\cite{he2015}, and random features, and report the results in \cref{tab:downstream}. Our approach outperforms the baselines in both tasks, indicating that we learn better spatial representations. 
Furthermore, our linear probe models show comparable performance against the supervised method which can be regarded as approximate upper bounds for our models, suggesting that our pretext tasks help learn a useful representation. Please see \supparxiv{supp.}{\cref{appendix:additional}} for experiments on FreeMusic~\cite{defferrard2016fma}.

\begin{figure}[b]
\centering
\upvspacefig
\includegraphics[width=\linewidth]{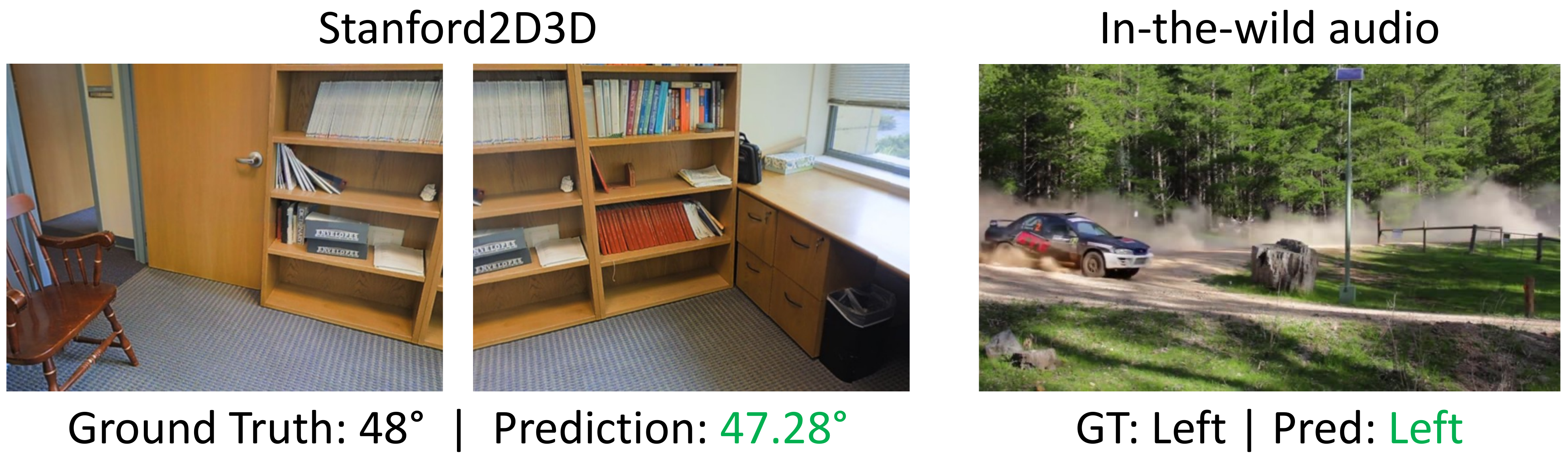}
    
\caption{
{\bf Qualitative results on real-world examples.} We show our predictions on Stanford2D3D~\cite{armeni2017joint} and In-the-wild audio~\cite{chen2022sound}. \textcolor{darkgreen}{Green} denotes accurate predictions.
}

\label{fig:real_world}
\end{figure}

\mypar{Emerging camera pose from audio prompting.}
To help understand the strong performance of our self-supervised features, we asked whether we could use the cross-view binauralization model {\em alone} to estimate camera rotation. Inspired by prompting in vision and language models~\cite{radford2021learning}, we obtain rough estimates of camera rotation by providing our model with carefully-provided inputs. 
Given a pair of images $(\bv_s, \bv_t)$, we create a synthetic binaural audio {\em prompt}, $\ba_s$. We then ask our model to generate the binaural sound $\hat{\ba}_t$ for the target viewpoint. By analyzing the IID cues in $\hat{\ba}_t$, we can estimate the model's implicitly predicted camera pose. To do this, we find the nearest neighbor of our generated audio $\hat{\ba}_t$, using a database of synthetically generated audio with known sound directions. Please refer to \supparxiv{supp.}{\cref{appendix:prompt}} for more details. Our method achieves the mean absolute error of $\mathbf{9.13^\circ}$ on HM3D-SS dataset where chance is $29.41^\circ$~(\cref{tab:pose_estimation}). Our approach can also generalize to Stanford2D3D~\cite{armeni2017joint} dataset, where we can achieve mean absolute error of $\mathbf{9.93^\circ}$~(\cref{tab:real_world}).

\subsection{Evaluating SLfM}
\label{sec:pose}
We evaluated our SLfM
models on the HM3D-SS dataset. We use the mean absolute error of angle in degrees as evaluation metrics.  To avoid sound field ambiguity, we filtered out samples with sound angles outside of $(-90^\circ, 90^\circ)$ for the evaluation set.
When training our SLfM models, we use our best-performing features, \ie, pretext tasks trained with 3 views.  We freeze learned audio and visual features and only train multi-layer perceptrons on top of them and evaluate them independently on each modality.
\begin{figure}[b]
\centering
\upvspacefig

\includegraphics[width=\linewidth]{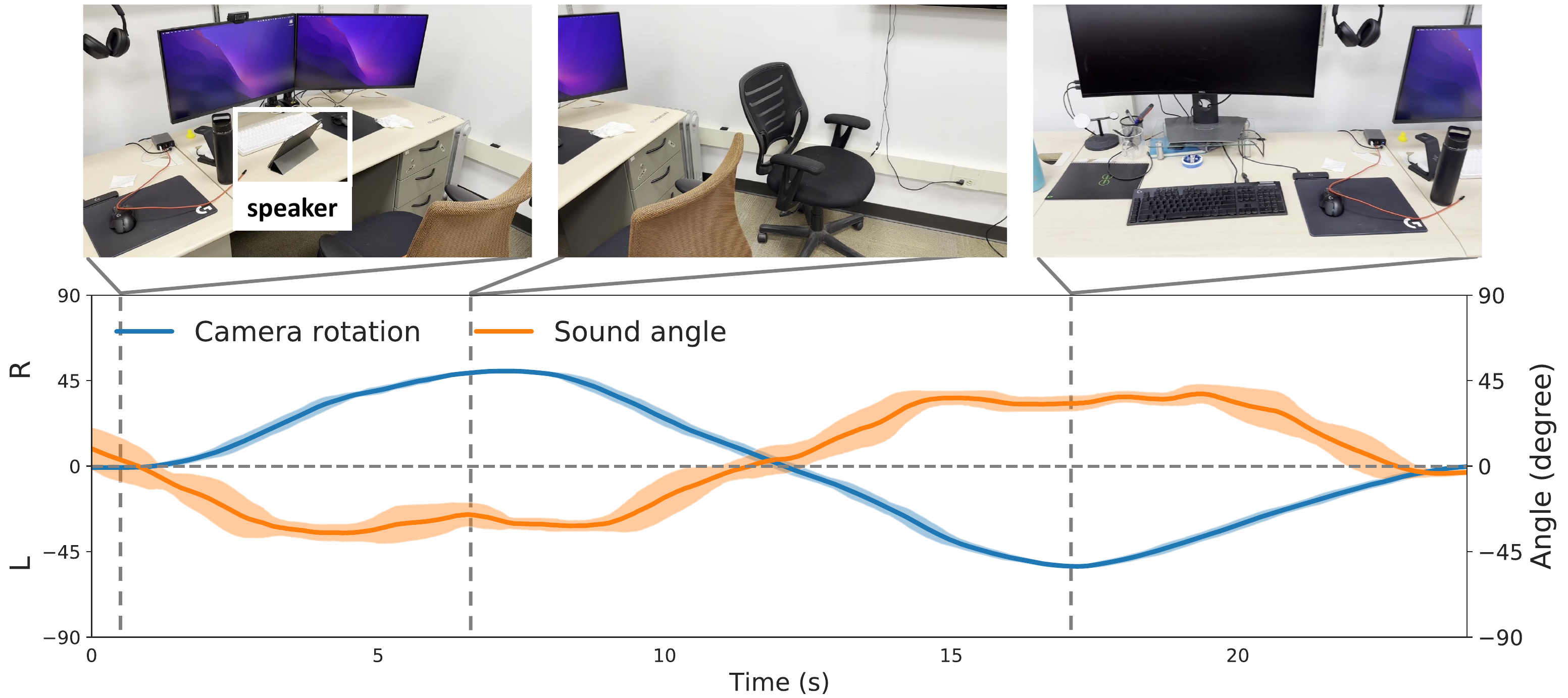}
    
\caption{
{\bf Real video.} We play music using a speaker and record a video using iPhone with a binaural microphone~(different from training examples). We show the predicted camera rotation and sound direction over time.
They change smoothly and match our camera motion. Please see our \href{https://ificl.github.io/SLfM}{project webpage} for video results.
 } 

\label{fig:demo}
\end{figure}

\begin{table}[t!]
\vspace{-1mm}
\upvspacefig
\begin{minipage}[b]{0.48\linewidth}
\centering
\resizebox{\columnwidth}{!}{
\begin{tabular}{lc}
    \toprule
    \multicolumn{2}{c}{In the wild audio~\cite{chen2022sound}} \\
    \midrule
     Model   & Acc~(\%)~$\uparrow$   \\ 
    \midrule
    GCC-PHAT~\cite{knapp1976Generalized}    &  77.2	   \\
    IID~\cite{chen2022sound}  &   75.4 \\
    IID~(Ours)  &   82.1 \\
    \cdashlinelr{1-2}
    MonoCLR~\cite{chen2022sound}   & {\bf 87.4}   \\
    StereoCRW~\cite{chen2022sound}  &  87.2  \\
    Ours--L2R  & 84.9 \\
    Ours  & 84.0 \\

    \bottomrule
\end{tabular}
}
\end{minipage}
\hfill
\begin{minipage}[b]{0.48\linewidth}
\centering
\raisebox{-0.18em}{
\resizebox{\columnwidth}{!}{
\begin{tabular}{lcc}
\toprule
\multicolumn{3}{c}{Stanford2D3D~\cite{armeni2017joint}} \\
    \midrule
\multirow{2}{*}{Model} &  \multicolumn{2}{c}{\shortstack[c]{Rot Err.~($^\circ$)~$\downarrow$}} \\
\cmidrule(lr){2-3}
  & Mean &  Med.  \\
\midrule
SIFT~\cite{lowe2004distinctive} & 16.4  & {\bf 0.06} \\
LoFTR~\cite{sun2021loftr}   &   6.10 & 1.13\\
Superglue~\cite{sarlin2020superglue} & 5.07 & 0.07  \\
Ours--Prompt & 9.93 &  9.02 \\
Ours--L2R & {\bf 1.12} &  0.71 \\
Ours & {1.14} &  0.67 \\
\bottomrule
\end{tabular}
}
}
\end{minipage}
\caption{{\bf Evaluation of our SLfM models on the real-world data.} We evaluate our audio localization model on the In-the-wild audio~\cite{chen2022sound}~(left) and our camera pose model on the Stanford2D3D~\cite{armeni2017joint}~(right). {\em Rot.} denote camera rotation. }
\label{tab:real_world}
\end{table}

\begin{figure*}[t]
    \centering
    \upvspacefig
    \vspace{-1mm}
    \includegraphics[width=\linewidth]{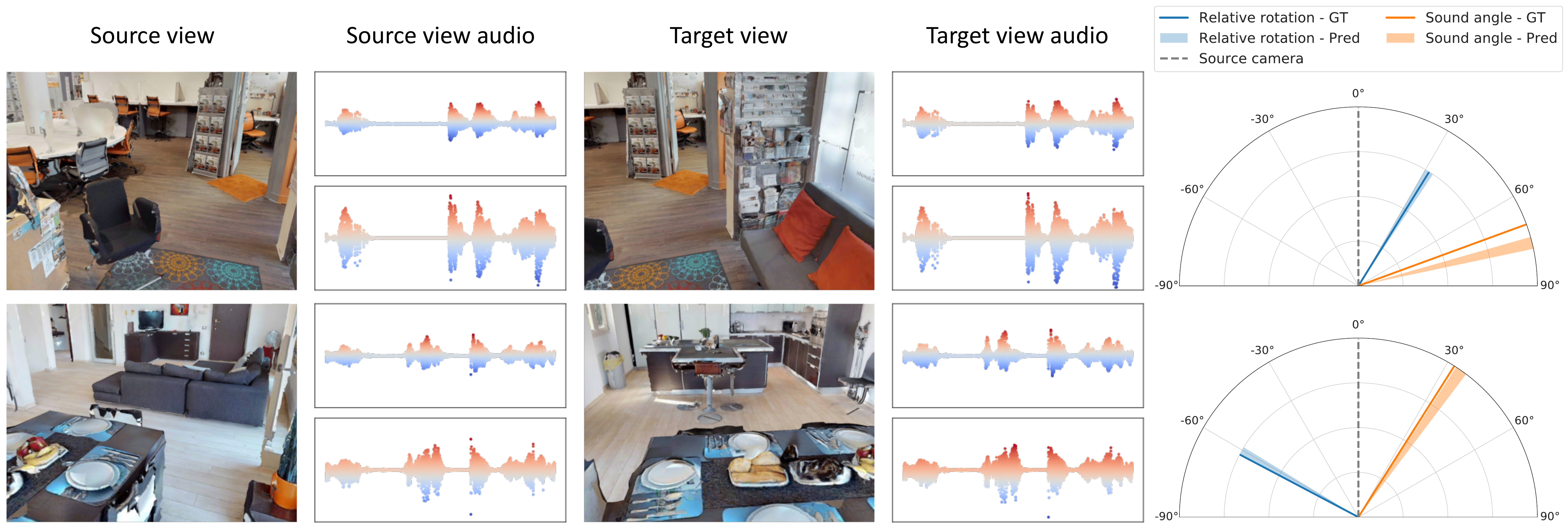}
    \caption{{\bf Qualitative results on HM3D-SS.} Given two images and their corresponding audio at the source viewpoints, our approach can individually predict the relative camera pose and localize sound locations  accurately. We visualize our camera rotation prediction with \textcolor{blue}{blue} bar and our sound angle prediction with \textcolor{orange}{orange} bar. To highlight the subtle differences in the waveforms, we color code the amplitude. Please see the \href{https://ificl.github.io/SLfM}{project webpage} for more video results.} 
    \label{fig:qualitative}
\end{figure*}

\mypar{Baselines and ablations.} 
For relative camera pose estimation, we compare our models with sparse feature matching using SIFT~\cite{lowe2004distinctive} and SuperGlue~\cite{sarlin2020superglue}, followed by rotation fitting.
We use those methods to detect key points, run Lowe's ratio test, and use RANSAC with five-point algorithm to recover the camera pose~\cite{lowe2004distinctive,Hartley04,nister2004efficient,fischler1981random}.
For the sound localization task, we compared with time delay estimation methods: the popular GCC-PHAT~\cite{knapp1976Generalized} and the recent self-supervised method StereoCRW~\cite{chen2022sound}. As far as we know, there are no other baselines that can estimate poses and sound source directions without labels. %

We also compared several variations of our method,
including {\bf Ours--Front}, where we filter out the samples with sound sources behind the viewers to remove binaural ambiguity during training, and an oracle model {\bf Ours--GTRot}, which uses ground-truth rotation angles instead.

\mypar{Results.} We show our results in \cref{tab:pose_estimation}. Our models can predict the azimuths of sound sources, obtaining strong performance without any labels. Without providing reflection invariance~(\cref{eq:per_geometric}), the model failed to learn reasonable geometric due to the audio ambiguity. Our self-supervised model achieves comparable performance against our oracle model~(Ours-GTRot) and supervised models, indicating we estimate camera poses and localize sound   accurately. 
We show some qualitative results on HM3D-SS in \cref{fig:qualitative} with LibriSpeech samples~\cite{panayotov2015librispeech}.

\mypar{Generalization to other datasets.}
We further demonstrate the generalization ability of our models by experimenting with out-of-distribution, real-world data. We evaluate our camera rotation model on Stanford2D3D~\cite{armeni2017joint} with real indoor RGB images (\cref{tab:real_world}). We obtain image pairs by cropping from panoramas. Although our model is trained on renderings of HM3D~\cite{ramakrishnan2021hm3d}, it obtains strong generalization ability. 
Compared with rotation estimation based on SIFT~\cite{lowe2004distinctive} and SuperGlue~\cite{sarlin2020superglue}, our model is significantly better on mean rotation error.
We also report median rotation error to be consistent with prior works~\cite{cai2021extreme,jin2021planar}.
However, SIFT~\cite{lowe2004distinctive} and Superglue~\cite{sarlin2020superglue} have a very low median error. This is likely due to the all-or-nothing nature of feature matching-based approaches, which either produce highly accurate predictions if the matches are correct (especially for ``easy'' cases with small amounts of rotation) or else produce gross errors.

We also evaluate our sound localization model on the in-the-wild binaural audio~\cite{chen2022sound}. We use binary accuracy as the metric for left-or-right direction classification accounting for the fact that microphone baselines are unknown for internet videos.
For an apples-to-apples comparison to prior work, we retrained our model using 0.51s length of audio for the binaural audio encoder $f_a(\cdot)$. 
As shown in \cref{tab:real_world}, our model obtains similar performance to StereoCRW, a state-of-the-art self-supervised time delay method, suggesting we have a strong capability for sound localization. We show qualitative results in \cref{fig:real_world}. We also perform both tasks on a self-recorded video~(\cref{fig:demo}) of a rotating camera and a binaural microphone. We show the mean and standard deviation of predictions in 1.0s windows.

\mypar{Handling ambiguity.} 
In binaural sound perception, there is a fundamental ambiguity that whether the sound is in front of or behind us. It leads to multiple solutions with equal loss in our model, mirroring sound sources and negating rotation angles with flipped $z$ axis~(\cref{fig:ambiguity}). These two solutions differ in that a visual rotation angle either indicates a clockwise or counterclockwise rotation. 
For evaluation, we convert ``backwards'' counterclockwise predictions to clockwise predictions by simply providing the model with pairs of input frames, then negating the model's outputs if the angle is the opposite of the expected direction\footnote{Similar to ambiguities in SfM where reconstructions can be reoriented such that the sky is in the positive $y$ direction~\cite{Hartley04,Hartley98e,agarwal2011building}.}.

\begin{figure}[b!]
\begin{minipage}{0.45\linewidth}
\upvspacefig
\input{floats/fig_ambiguity}
\end{minipage}
\hfill
\begin{minipage}{0.45\linewidth}
\upvspacefig
\vspace{-4.0mm}
\input{floats/fig_reverb_pose}
\end{minipage}

\end{figure}

\subsection{Generalization to more complex scenarios}
\label{sec:complex_scene}
We investigate whether our approach generalizes to more complex scenarios, such as with multiple sound sources or when translation is included in camera motion.

\begin{table}[t!]
\centering
\upvspacefig
\resizebox{0.9\columnwidth}{!}{
\begin{tabular}{lccccc}
\toprule
\multirow{2}{*}{Model} & \multicolumn{2}{c}{\shortstack[c]{Multi-source\\ Acc~(\%)~$\uparrow$}} &  & \multicolumn{2}{c}{\shortstack[c]{Small trans.\\ Acc~(\%)~$\uparrow$}}  \\
\cmidrule(lr){2-3}
\cmidrule(lr){5-6}
 &  Aud. & Rot. &  & Aud.  & Rot.           \\
\midrule

Random feature  &  4.5 & 4.7  &  & 5.7 & 3.8 \\
ImageNet~\cite{he2016}+Random &  -- &  56.3   &   & -- &  23.0  \\
RotNCE~\cite{francl2022modeling} &  35.8    &   -- &   &  --  & -- \\
AVSA~\cite{morgado2020learning} &  55.9 &  6.7  &   & 65.4 & 6.1 \\
\cdashlinelr{1-6}
Ours~(3 views) & {\bf 59.2} & {\bf 77.9} &   & {\bf 72.2} & {\bf 49.6}\\
\cmidrule(lr){1-6}
Supervised  &  74.5 & 95.8 &  & 81.6 & 63.5\\
\bottomrule
\end{tabular}
}
\caption{{\bf Generalization to more complex scenarios.} We evaluate the audio and visual features learned from more complex scenarios via linear probing. {\em Aud.} and {\em Rot.} denote audio localization and camera rotation respectively. }
\label{tab:complex}
\end{table}

\mypar{Multiple sound sources.} 
We evaluate our representations and SLfM models on more complex scenes containing multiple sound sources. We train our models with two source sources placed in the scenes. One of them is dominant and our sound localization target. 
We report the results in \cref{tab:complex} and \cref{tab:complex_pose}. Our model learns better representations than baselines and achieves accurate predictions of the azimuth of the source and camera pose even in challenging scenarios with multiple sound sources.

\begin{table}[t!]
\small
\centering
\resizebox{0.82\columnwidth}{!}{
\begin{tabular}{lccccc}
\toprule
\multirow{2}{*}{Model} & \multicolumn{2}{c}{\shortstack[c]{Multi-source\\MAE~($^\circ$)~$\downarrow$}}  &   &  \multicolumn{2}{c}{\shortstack[c]{Small trans.\\ MAE~($^\circ$)~$\downarrow$}}  \\
\cmidrule(lr){2-3}
\cmidrule(lr){5-6}
&  Aud. & Rot. &   & Aud.  & Rot.           \\
\midrule

Chance &  40.46  & 29.41 &   & 40.28   & 29.41  \\
SIFT~\cite{lowe2004distinctive} &  -- & 12.2 &   &  --  & 12.2 \\
\cdashlinelr{1-6}
Ours &  {\bf 7.67}   & {\bf 0.71}  &  &  {\bf 6.28} & {\bf 1.04} \\
\cmidrule(lr){1-6}
Ours--GTRot &  5.81  & --  &  & 4.24 & --\\
\cmidrule(lr){1-6}
Superglue~\cite{sarlin2020superglue} & -- & 2.47 &   & --  & 2.47 \\
Supervised  &  3.60 & 0.46 &   & 1.71 & 0.46 \\
\bottomrule
\end{tabular}
}

\caption{{\bf SLfM results on more complex scenarios.} We evaluate our model on the version of HM3D-SS with two sound sources. We also evaluate our model trained with small translations on rotation-only examples. }\vspace{-3mm}
\label{tab:complex_pose}
\end{table}

\mypar{Translation in camera motions.} 
To study how small translations in the camera motion could affect our models,  we generate 50K pairs of audio-visual data with both rotation and translation change, limiting the uniformly sampled translation to 0.5 meters. We train our models with LibriSpeech samples~\cite{panayotov2015librispeech}. 
For our linear probing evaluation, we measure the ability of features to handle complex examples by testing on the dataset that has translation. 
For our SLfM model, we study our ability to learn from noisy data, thus we evaluate it on the rotation-only examples.
As the results are shown in \cref{tab:complex} and \cref{tab:complex_pose}, we successfully learn useful features and obtain accurate rotation and sound direction predictions despite the presence of translation. Since we jointly learn the audio and visual representations, it can negatively impact the learning of one modality when another one becomes harder.

\subsection{Ablation study}
\vspace{3mm}

\mypar{Robustness to reverberation.} 
We study how our SLfM models perform under the different reverberation configurations. We used SoundSpaces~\cite{chen2022soundspaces} to create audio with average reverberation $\text{RT}_{60} \in \{0.1, 0.2, 0.3, 0.4, 0.5\}$ while keeping the visual signals the same.
We train our model on each setting. As shown in \cref{fig:reverb_pose}, our performance decreases as the level of reverberation increases, where audio becomes more challenging during both training and testing. Please see \supparxiv{supp.}{\cref{appendix:ablation}} for the study on representations.

\mypar{Losses for SLfM.} 
We study the necessity of our proposed loss functions in \cref{tab:loss_ablation}.
Our models fail to learn accurate pose estimation without binaural loss or symmetric loss.
It highlights the crucial role of these losses.

\begin{table}[t!]
\centering
\upvspacefig
\resizebox{1\columnwidth}{!}{
\begin{tabular}{lccccc}
\toprule
\multirow{2}{*}{Model} & \multicolumn{3}{c}{Losses} & Audio angle & Camera angle\\
\cmidrule(lr){2-4}
 & $\mathcal{L}_{\mathrm{geo}}$ & $\mathcal{L}_{\mathrm{binaural}}$ & $\mathcal{L}_{\mathrm{sym}}$ & MAE~($^\circ$)~$\downarrow$ &  MAE~($^\circ$)~$\downarrow$ \\
\midrule
\multirow{4}{*}{\shortstack[c]{Ours}}  &  \cmark &    &  & 37.60 & 29.20 \\
  &  \cmark &    &  \cmark & 37.52 & 29.17 \\
 &  \cmark &   \cmark &  & 3.58 & 6.99 \\

   & \cmark  &  \cmark  & \cmark & {\bf 3.17} & {\bf 0.77} \\
\bottomrule
\end{tabular}
}
\caption{{\bf Ablation experiments on our SLfM losses.} We evaluate our SLfM models with different combinations of losses. }
\vspace{-2mm}
\label{tab:loss_ablation}
\end{table}

\section{Conclusion}
In this paper, we proposed the {\em sound localization from motion} (SLfM) problem, and provided a self-supervised method for solving it. We also presented a method for learning audio-visual features that convey sound directions and camera rotation, which we show are well-suited to solving the SLfM task. 
Despite learning our models solely from unlabeled audio-visual data, we obtain strong performance on a variety of benchmarks, including rotation estimation on the Stanford2D3D~\cite{armeni2017joint} dataset and ``in the wild'' sound direction estimation~\cite{chen2022sound}. Our results suggest that the subtle correlations between sights and binaural sounds that result from rotational motion provide a useful (and previously unused) learning signal. We see our work as opening new directions in self-supervised geometry estimation and feature learning that use sound as a complementary source of supervision. We will release code, data, and models upon acceptance.

\mypar{Limitations and Broader Impacts.} 
Our work has several limitations. 
First, while we evaluate our models on real images and sounds, we train on data from simulators, due to a lack of available relevant data. We note that this is common practice in visual 3D reconstruction~\cite{jin2021planar,wang2022neuris,lin2022neurmips}. %
Second, we assume that the 3D scene and sound sources are stationary.
Third, we do not evaluate our model on extreme viewpoint changes~\cite{ma2022virtual}, which requires reasoning about images that have little or no overlap.

\mypar{Acknowledgements.}
We would like to thank Ang Cao, Tiange Luo, and Linyi Jin for their helpful discussion. We thank Changan Chen for his help with SoundSpaces 2.0.  This work was funded in part by DARPA Semafor and Sony. The views, opinions and/or findings expressed are those of the authors and should not be interpreted as representing the official views or policies of the Department of Defense or the U.S. Government. %

{\small
\bibliographystyle{ieee_fullname}
\bibliography{audiosfm}
}

\clearpage
\appendix
\supparxiv{
\setcounter{page}{1}
\twocolumn[{%
\renewcommand\twocolumn[1][]{#1}%
\begin{center}
    \vspace{-1.0em}
    {\bf \large Supplementary Material for Sound Localization from Motion:\\ Jointly Learning Sound Direction and Camera Rotation}
    \vspace{1.0em}
\end{center}

}]

}{}
\renewcommand{\thesection}{A.\arabic{section}}
\setcounter{section}{0}

\section{Camera pose from audio prompting}
\label{appendix:prompt}
\begin{figure}[h!]
\centering
\includegraphics[width=\linewidth]{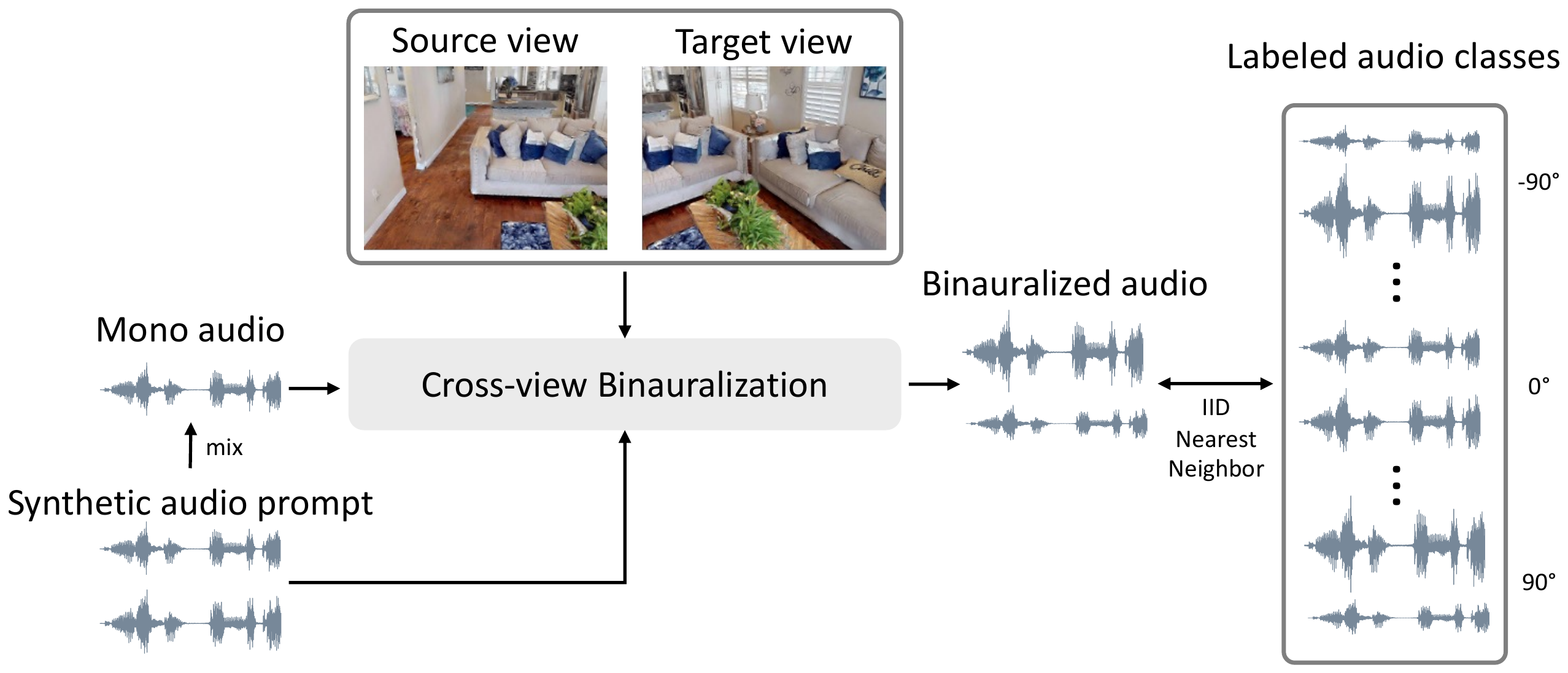}
    
\caption{
{\bf Estimating camera pose from audio prompting.} We estimate camera rotation by providing our cross-view binauralization model with synthetically generated audio prompts. Given the sound that it predicts, we infer the camera angle. We do this by finding the nearest neighbor (using IID cues) to a database of synthetic sounds, each paired with their corresponding angle.
}

\label{fig:prompt}
\end{figure}

We illustrate our prompting idea in \cref{fig:prompt}.
To create our audio prompts, we simulate 181 binaural RIRs at different angles from $[-90^\circ, 90^\circ]$ without reverberation using SoundSpaces~\cite{chen2022soundspaces} and render with audio signals from LibriSpeech~\cite{panayotov2015librispeech}. We use the sound with an angle of $0^\circ$ as the input prompt $\ba_s$~(the source view audio) and mix it into mono audio as the input at the target viewpoint. We calculate the interaural intensity difference (IID) cues for the audio prompts $\ba_i$ and generated audio $\hat{a}_t$. We use L1 distance between IID cues to find the nearest neighbors:
\begin{equation}
     \argmin_{\bA_i} \left\vert \log_{10}\frac{\hat{\bA}^L_t}{\hat{\bA}^R_t} - \log_{10}\frac{\bA^L_i}{\bA^R_i}  \right\vert,
    \label{eq:prompt}
\end{equation}
where $\bA_i=\text{STFT}(\ba_i)$. We use ground truth annotations of sound directions from the nearest prompts to predict the camera rotation angles. We first obtain rotation prediction votes from 1024 audio prompts and use a RANSAC-like mode estimation~\cite{fischler1981random,chen2022sound} to get the final prediction.

\section{Additional experimental results}
\label{appendix:additional}
\vspace{3mm}
\mypar{Evaluating pretext task.}
We also evaluate the performance of our model on the pretext task, which involves binauralizing sound at a novel microphone pose using sound from a different viewpoint and visual cues from both views as references.  We use the STFT distance between the predicted and ground-truth spectrogram to measure the audio reconstruction performance. As the results are shown in \cref{tab:pretext}, our model that incorporates both visual and audio features as input performs the best and is comparable to the model that receives ground truth rotation angles as inputs. 
This suggests that our model effectively uses the spatial information in both visual and audio signals to solve binauralization tasks, and encourages the network to learn useful representations. Moreover, the results show that training with more viewpoints improves the performance of the pretext task.

\begin{table}[!ht]
\centering
\resizebox{0.9\columnwidth}{!}{
\begin{tabular}{clw{c}{2em}cc}
\toprule
& \multirow{2}{*}{Model} & \multicolumn{2}{c}{Input features} & \multirow{2}{*}{STFT distance~$\downarrow$} \\
\cmidrule(lr){3-4}
&   & \multicolumn{1}{c}{$\mathcal{V}$} & \multicolumn{1}{c}{$\mathcal{A}$} &  \\
\midrule

\parbox[t]{2mm}{\multirow{7}{*}{\rotatebox[origin=c]{90}{\shortstack[c]{Mono2Binaural }}}} & Random  &  \cmark &   \cmark & 0.368 \\
 \cdashlinelr{2-5}
& \multirow{4}{*}{\shortstack[c]{Ours~(2 views)}}  &   &    & 0.206 \\
 &  &   \cmark &     & 0.207 \\
 &  &   &    \cmark  & 0.161 \\
 &  &   \cmark &    \cmark &  0.130 \\
 \cdashlinelr{2-5}
 & Ours--GTRot &   &    \cmark &   0.131 \\
  & Ours~(3 views) & \cmark  &    \cmark &   {\bf 0.125} \\

\bottomrule
\end{tabular}
}
\caption{{\bf Reconstruct performance of cross-view binauralization pretext task.} We report the STFT distance performance of variants of our models with different input features on HM3D-SS dataset with LibriSpeech samples~\cite{panayotov2015librispeech}. $\mathcal{V}$ and $A$ mean visual and audio features, respectively.}
\label{tab:pretext}
\end{table}

\mypar{Experiment on FreeMusic.} We report the performance of downstream tasks with learned representations on the HM3D-SS dataset with FreeMusic~\cite{defferrard2016fma} samples in \cref{tab:downstream_fma}. We outperform baselines and learn a useful representation.

\begin{table}[h!]
\centering

\resizebox{0.95\columnwidth}{!}{
\begin{tabular}{clcc}
\toprule
& \multirow{2}{*}{Model} & \multirow{2}{*}{\shortstack[c]{Audio Loc. \\ Acc~(\%)~$\uparrow$}} & \multirow{2}{*}{\shortstack[c]{Camera Rot. \\ Acc~(\%)~$\uparrow$}} \\
&   &  &              \\
\midrule

\parbox[t]{1mm}{\multirow{8}{*}{\rotatebox[origin=c]{90}{\shortstack[c]{FreeMusic}}}} 
& Random feature &  6.0 & 4.7\\
& ImageNet~\cite{he2016}+Random &  --  & 56.3 \\
& RotNCE~\cite{francl2022modeling} &  46.3  & -- \\
& AVSA~\cite{morgado2020learning} &  66.5  & 6.7 \\
\cdashlinelr{2-4}
& Ours--L2R~(3 views) & {\bf 72.0}  & 76.5\\ 
& Ours~(2 views) &  67.5  & 76.2\\
& Ours~(3 views) &  67.5  & {\bf 81.1}\\
\cmidrule(lr){2-4}
& Supervised & 77.1 & 95.8 \\
\bottomrule
\end{tabular}
}

\caption{{\bf Downstream task performance on HM3D-SS dataset with FreeMusic~\cite{defferrard2016fma} samples.} We report linear probe performance on the audio localization and camera rotation downstream tasks.  }

\label{tab:downstream_fma}
\end{table}

\mypar{SLfM without pretraining.} We further demonstrate the important role of the features learned from our cross-view binaural pretext task by training our SLfM model with random features. We show results in \cref{tab-appendix:random_pose}. We can see that the models perform better using our feature representations, which emphasizes the significance of our pretext task. Our SLfM model finetuned from random features achieves accurate predictions, highlighting that our proposed method successfully leverages the geometrically consistent changes between visual and audio signals.

\begin{table}[h!]
\centering
\vspace{-2mm}
\resizebox{1\columnwidth}{!}{
\begin{tabular}{llcc}
\toprule
\multirow{2}{*}{Model} & \multirow{2}{*}{Init. feature}  &  \multirow{2}{*}{\shortstack[c]{Audio angle\\MAE~($^\circ$)~$\downarrow$}} & \multirow{2}{*}{\shortstack[c]{Camera angle\\MAE~($^\circ$)~$\downarrow$}} \\
&   & &       \\
\midrule
Ours &  Random~(freeze)  &  36.51    & 29.26 \\
Ours &  Random~(finetune)  &  3.92    & 1.32 \\
Ours &   M2B ~(freeze) &{3.17}   & { 0.77} \\
Ours &   M2B ~(finetune) &{\bf 2.77}   & {\bf 0.76} \\

\bottomrule
\end{tabular}
}

\caption{{\bf SLfM results with different features.} We evaluate our SLfM models trained with different feature initialization on HM3D-SS.}
\label{tab-appendix:random_pose}
\end{table}

\section{Ablation study}
\label{appendix:ablation}
\vspace{3mm}

\mypar{Audio prediction network.} We study how audio prediction architectures will influence representation learning from our proposed pretext task. We adapt the U-Net architecture with cross-attention modules for conditional feature inputs~\cite{rombach2022high,vaswani2017attention} and compare the pretext and downstream performance with U-Net~\cite{gao20192} we used for our main experiments. We train our models on the HM3D-SS dataset with a single sound source presented in the scenes and use LibriSpeech signals~\cite{panayotov2015librispeech}. We report results in \cref{tab:unet}. Interestingly, we found that ATTN U-Net can reconstruct better sounds for the pretext task while it does not learn the features as well as the 2.5D U-Net~\cite{gao20192}. We hypothesize that a more complex network may transfer the representation learning inside of the prediction networks rather than the feature extractors. 
\begin{table}[ht!]
\centering
\resizebox{1\columnwidth}{!}{
\begin{tabular}{lccc}
\toprule
\multirow{2}{*}{Model} & Pretext~$\downarrow$ &  \multicolumn{2}{c}{\shortstack[c]{Downstream Acc~(\%)~$\uparrow$}} \\
\cmidrule(lr){2-2}
\cmidrule(lr){3-4}
 & STFT Dist. & AudLoc. &  CamRot.  \\
\midrule
ATTN U-Net~\cite{rombach2022high,vaswani2017attention} &  {\bf 0.128} & 68.0 & 75.3 \\
2.5D U-Net~\cite{gao20192} &  0.130 & {\bf 74.5} & {\bf 80.0} \\
\bottomrule
\end{tabular}
}
\caption{{\bf Audio prediction model ablation study.} We evaluate both pretext and downstream performance on the HM3D-SS with LibriSpeech samples~\cite{panayotov2015librispeech}.}
\label{tab:unet}
\end{table}

\renewcommand{\columnsep}{10pt}
\begin{wraptable}{r}{0.4\linewidth}
\vspace{-1.0em}
\captionsetup{type=figure}
\centering

\includegraphics[width=1.0\linewidth]{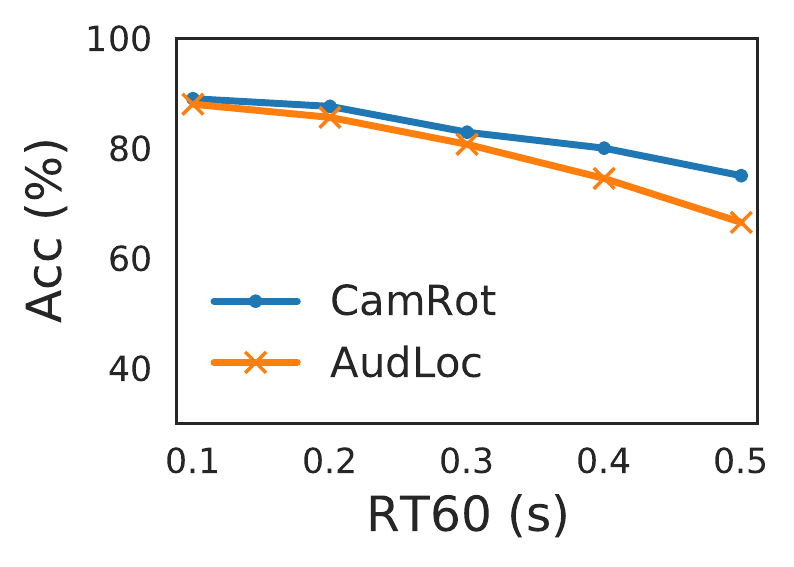}
    
\caption{
{\bf Robustness to reverberation.} We study the effect of reverberation on our pretext model. Chance performance is 1.5\%.
} 
\label{fig:reverb_pretext}

\vspace{-1.0em}
\end{wraptable} %
\mypar{Robustness to reverberation.} We also evaluate our representation under the influence of reverberation. We report linear probe performance on downstream tasks with average reverberation $\text{RT}_{60} \in \{0.1, 0.2, 0.3, 0.4, 0.5\}$. 
As shown in \cref{fig:reverb_pretext}, the results indicate a decrease in downstream performances as the level of reverberation increases, where audio becomes more challenging during both training and testing.

\begin{figure}[t!]
\centering

\includegraphics[width=0.88\linewidth]{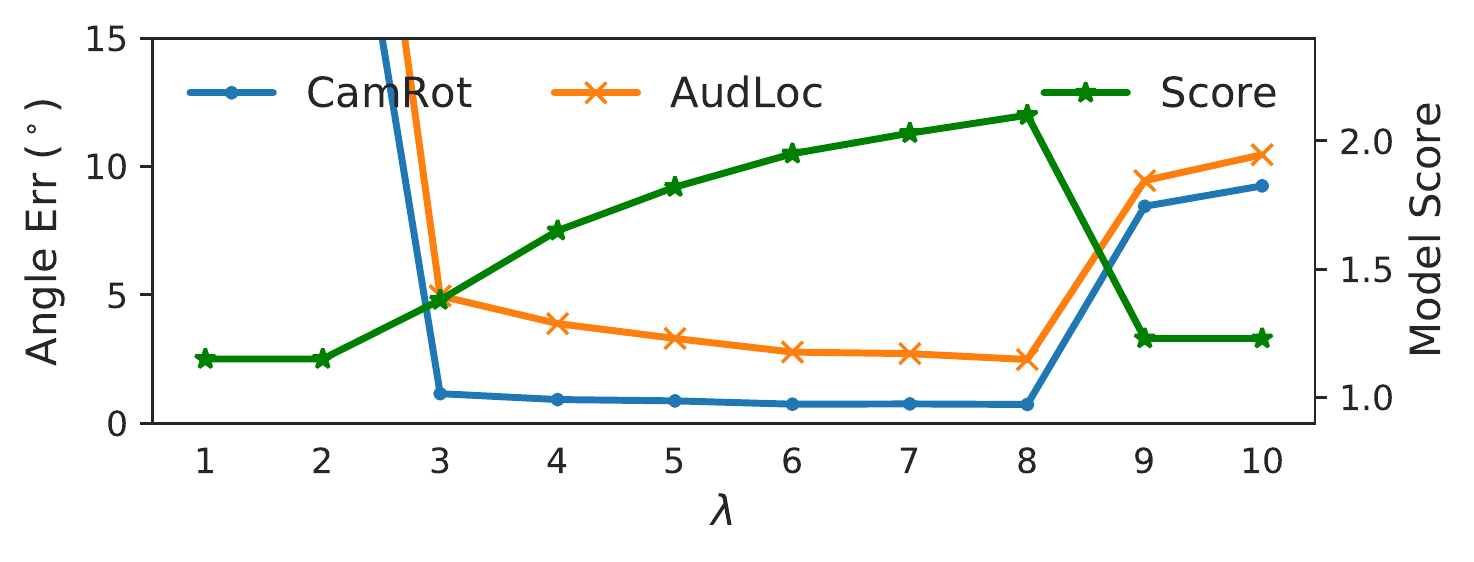}

\vspace{-2mm}
\caption{
{\bf Hyparameter search.} We experiment with $\lambda$ from 1 to 10 and monitor the scores of models.
} 
\vspace{-3mm}

\label{fig:hparam_search}
\end{figure}

\mypar{Weights of geometric loss.} 
We assign appropriate weights for the geometric loss~(\cref{eq:geometric}) to avoid it from dominating the optimization. 
In our approach, we search $\lambda$ from 1 to 10 during training, and we  select models weights using a metric by calculating $1 / (100 \cdot \mathcal{L}_{\mathrm{geo}} + \mathcal{L}_{\mathrm{binaural}} + \mathcal{L}_{\mathrm{sym}})$ during validation. We show the search experiment in \cref{fig:hparam_search}. The performance is relatively stable when $\lambda \in [3,8]$. We select $\lambda=5 \text{ or } 3$  in the main paper, please see \cref{appendix:implement} for details.

\section{Implementation details}
\label{appendix:implement}
\vspace{3mm}
\mypar{SLfM model.} We use separate multi-layer perceptrons $g_v$ and $g_a$~(\ie, FC~($512\rightarrow 256$)--ReLU--FC~($256\rightarrow 1$) layers) to predict scalar rotation and sound angles.

\mypar{Hyperparameters.}
For all experiments, we re-sample the audio to 16kHz and use 2.55s audio for the binauralization task. For pretext training, we use the AdamW optimizer~\cite{kingma2015adam,loshchilov2017decoupled} with a learning rate of $10^{-4}$, a cosine decay learning rate scheduler, a batch size of 96, and early stopping. During downstream tasks, we change the learning rate to $10^{-3}$ for linear probing experiments. To train our self-supervised pose estimation model, we set the weights $\lambda$ of geometric loss to be $5$ and weights of binaural and symmetric losses to be $1$. For more complex scenarios~(\cref{sec:complex_scene}), we set the weights $\lambda$ as $3$ to avoid the geometric loss from dominating.

\mypar{IID cues.} We describe our implementation of predicting sound on the left or right using IID cues in detail here: we first compute the magnitude spectrogram $|\bA|$ from the binaural waveform $\ba$ and sum the magnitude over the frequency axis. Next, we calculate the log ratio between the left and right channels for each time frame. After this, we take the sign of log ratios and convert them into either +1 or -1. We sum over the votes and take the sign of it for final outputs.

\mypar{Dataset.} Due to the fact that SoundSpaces 2.0~\cite{chen2022soundspaces} does not support material configuration for HM3D~\cite{ramakrishnan2021hm3d} at the current time, we obtain binaural RIRs with different reverberation levels by scaling the indirect RIRs and add them up with direct RIRs. 
We render binaural sounds with random audio samples as augmentation during training.

\end{document}